\documentclass{article}

\usepackage{arxiv}

\usepackage[utf8]{inputenc}
\usepackage[T1]{fontenc}
\usepackage{hyperref}
\usepackage{url}
\usepackage{booktabs}
\usepackage{amsmath}
\usepackage{amsfonts}
\usepackage{amssymb}
\usepackage{mathtools}
\usepackage{nicefrac}
\usepackage{microtype}
\usepackage{xcolor}
\usepackage{graphicx}
\usepackage{caption}

\graphicspath{{plots/}{figures/}}

\title{Edge-aware Decoding for Neural Asymmetric Routing}

\author{
Li Liang \\
Sun Yat-Sen University \\
\texttt{liangli23@mail2.sysu.edu.cn}
\And
Jinbiao Chen \\
Department of Industrial Systems Engineering and Management, \\
National University of Singapore \\
\texttt{bill.cjb@nus.edu.sg}
\And
Zizhen Zhang \\
Sun Yat-Sen University \\
\texttt{zhangzizhen@gmail.com}
}

\begin{document}

\maketitle

\begin{abstract}
Neural asymmetric routing models increasingly encode directionality through
matrix representations and asymmetry-aware attention. The final routing action,
however, is not a node in isolation but a directed transition chosen under the
current partial route. This creates a representation--decision mismatch:
pairwise cost information may be encoded upstream while the final candidate
logit is still largely parameterized as context--node compatibility. We propose
a decoder-design principle for neural asymmetric routing: the final score should
explicitly expose transition-level quantities suggested by the problem's
cost-to-go structure. We instantiate this principle with an edge-aware decoder
that adds candidate-specific terms for the current directed edge,
return-to-start closure, and static lightweight lookahead, while keeping the
representation backbone fixed. On a controlled SVD/Sinkhorn asymmetric backbone,
the decoder improves over the RADAR reference when trained on ATSP-100 and
evaluated zero-shot on ATSP-100/200/500/1000, reducing the ATSP-1000 gap from
$4.13\%$ to $2.73\%$. On ACVRP, the same score-level modification shows the same
qualitative trend under a richer routing state. ATSP ablations and
directed-transition diagnostics sharpen the mechanism: the strongest evidence
concerns sensitivity to the current directed edge, while closure and static
lookahead act as heuristic continuation cues. The results support a mechanism
study: a key decoder-side signal in neural
asymmetric routing is decision-time exposure of transition-level edge
information.
\end{abstract}

\section{Introduction}
\label{sec:introduction}

Asymmetric routing is challenging not only because travel costs are directional, but also because the relevant direction shifts with the construction state. One-way streets, turn penalties, depot return, and route continuation all render the transition \(i \rightarrow j\) fundamentally different from \(j \rightarrow i\). Neural combinatorial optimization has advanced rapidly on routing problems \citep{kool2019attention,kwon2020pomo}, yet asymmetric routing still exposes a basic tension: the solver must reason over directed pairwise costs while making a sequence of state-dependent decisions.

Recent asymmetric routing neural solvers address this tension mainly on the representation side. Matrix-based models such as MatNet operate directly on cost matrices rather than only on Euclidean coordinates \citep{kwon2021matnet}. RADAR pushes this direction further by combining SVD-based initialization with Sinkhorn-normalized attention, so that asymmetry enters both the initial representation and encoder interactions \citep{yi2026radar}. In this paper, we investigate whether such representation-side asymmetry is explicitly exploited when the model commits to the next action.

This question is broader than any single asymmetric backbone. A model can expose
both \(D_{ij}\) and \(D_{ji}\) to its encoder, yet still score the next action
through a mostly node-centric final logit. We call this a
\emph{representation--decision mismatch}: directed matrix information is encoded
upstream, while the final action score does not explicitly expose the current
directed transition. This is not a claim that existing decoders are state-free:
they condition on quantities such as the previous node, start node, depot state,
and feasibility masks. The narrower issue is that the final candidate logit is
often parameterized as node compatibility rather than explicit transition value.

The mismatch has a structural explanation. In asymmetric routing, the action is a directed transition under the current mask and partial route. The decoder therefore needs to approximate a state-conditioned transition value, not a context-free notion of node quality. For the Asymmetric Traveling Salesman Problem (ATSP), this perspective is naturally expressed through the Bellman form
\(
Q^*(i,j,U)=D_{ij}+V(j,U\setminus\{j\})
\),
which depends on the immediate edge cost and the continuation value after choosing $j$; the return-to-start cost appears through the terminal boundary condition. A node-centric score can recover such information only indirectly from embeddings. That is plausible in principle, but it is a weak inductive bias for asymmetric matrix routing.

We therefore advocate a decoder-side design principle: the final score of a constructive asymmetric-routing solver should expose transition-level quantities suggested by the problem's cost-to-go structure. We instantiate this principle as a score-level edge-aware decoder with explicit terms for the current directed transition, closure, and static lightweight lookahead. Crucially, we keep the SVD/Sinkhorn asymmetric representation backbone fixed, isolating the effect of decision-time edge exposure from representation-side changes.

On ATSP, which isolates asymmetric edge selection in its purest form, our decoder-side edge-aware scoring consistently improves over the RADAR reference across ATSP-100/200/500/1000 under the same benchmark setting. The Asymmetric Capacitated Vehicle Routing Problem (ACVRP) demonstrates the same qualitative trend under a richer routing state. ATSP decoder ablations and directed-transition diagnostics then localize the mechanism: the strongest empirical signal is the sensitivity to the current directed edge, while closure and static lookahead act as useful continuation cues rather than independent explanations.

This work makes four contributions. First, we identify representation–decision mismatch as a concrete decoder-side bottleneck in neural asymmetric routing. Second, we employ a Bellman-style transition-value viewpoint to motivate a general decoder design principle: final logits should expose action-level quantities including the immediate directed edge cost, return-to-start closure, and lightweight continuation cues. Third, we instantiate this principle as an edge-aware decoder on a fixed SVD/Sinkhorn asymmetric backbone, establishing new state-of-the-art results for asymmetric routing neural solvers. Fourth, we provide ablations and counterfactual diagnostics that disentangle immediate directed-edge sensitivity from broader continuation effects. We report the additional inference cost explicitly in the experiments and discuss it as a limitation.

\section{Related Work}
\label{sec:related-work}

\textbf{Matrix-encoded neural solvers for ATSP.}
ATSP is naturally specified by a directed cost matrix, so neural solvers must move beyond the coordinate-only view common in symmetric Euclidean routing. \citet{kool2019attention} and \citet{kwon2020pomo} established the attention-based encoder--decoder paradigm for autoregressive construction, while matrix-encoded ATSP is more directly represented by MatNet. \citet{kwon2021matnet} introduced a neural architecture for matrix-style relationship data and applied it to ATSP and FFSP, showing that attention-based construction can operate directly on pairwise cost matrices. \citet{pan2025unico} further broadened this view by reducing multiple combinatorial optimization problems to matrix-encoded general TSP, including settings with non-metric, asymmetric, or discrete distances. These works extend neural routing from coordinate-based instances to matrix-input problems and establish the modeling foundation for matrix-encoded ATSP. Their main emphasis, however, is on how to encode cost matrices and adapt constructive policies to such inputs, leaving the role of matrix entries during the final action-scoring step less explicitly studied.

\textbf{Representation and encoder design for asymmetric routing.}
Recent work on ATSP and asymmetric VRPs has largely strengthened representation and backbone design. \citet{yi2026radar} proposed RADAR, which uses SVD-based initialization to capture static source--target asymmetry and Sinkhorn-normalized attention to model dynamic asymmetric interactions. \citet{son2025rrnco} studied real-world routing with asymmetric distance and duration matrices, using contextual gating, adaptive fusion or bias, and edge/context signals such as distance, duration, and direction or angle. \citet{zhou2024icam} addressed instance-conditioned adaptation for cross-scale generalization rather than real-world asymmetric routing specifically. Other large-scale or unified routing systems improve global--local construction, transferable local policies, divide-and-conquer inference, or foundation-style coverage over VRP variants \citep{ye2024glop,gao2024elg,zheng2024udc,berto2024routefinder}. Together, these methods improve representation-side asymmetry, real-world adaptation, and scale generalization. They mainly answer how an encoder or backbone should represent and process asymmetric matrices. However, an asymmetry-aware representation does not by itself specify how the decoder should use directed information when assigning the final score to a feasible next node. Whether the asymmetry captured by the backbone is explicitly available to, and effectively used by, the final decoder logits remains less explored.

\textbf{Decoder design and transition-level scoring.}
Decoder design has also emerged as an important factor in neural routing. \citet{luo2023lehd} used a light encoder and a heavy decoder so that decoding can dynamically capture relationships among available nodes and improve large-scale generalization. \citet{huang2025reld} observed that light decoders rely heavily on static encoder embeddings and may struggle to extract context-specific decision information, and therefore increased decoder capacity with identity mapping and feed-forward layers. These works show that the decoder matters, but they mainly improve general decoder capacity. A related line studies score-level structural cues: \citet{ying2021graphormer} demonstrated the usefulness of pairwise structural bias in graph Transformers, \citet{wang2024dara} reshaped routing attention scores with Euclidean distance information, \citet{tran2025llm} injected externally generated structural cues into neural solver attention scores, and \citet{son2025rrnco} used adaptive bias to incorporate real-world edge/context signals. These methods support the usefulness of structural cues at the score level, but they typically target encoder attention, generic attention reshaping, Euclidean proximity, real-world contextual adaptation, or external cues. In contrast, ATSP requires the decoder to score a directed transition under the current partial route. From a Bellman-style action-value perspective, choosing \(j\) from current node \(i\) should reflect not only the immediate directed cost \(D_{ij}\), but also lightweight proxies for future route value, such as return-to-start closure and continuation cost. This motivates decoder-side edge-aware scoring, moving from representation-side asymmetry to decision-time transition scoring.

\section{Methodology}

\subsection{Problem Formulation and Optimal Decision Rule}

We consider the Asymmetric Traveling Salesman Problem (ATSP) specified by a directed cost matrix $D\in\mathbb{R}^{n\times n}$, where $D_{ij}\neq D_{ji}$ in general. A solution is a permutation $\pi=(\pi_1,\dots,\pi_n)$ minimizing the tour cost $\sum_{t=1}^{n-1}D_{\pi_t,\pi_{t+1}}+D_{\pi_n,\pi_1}$. The autoregressive construction policy selects the next node step by step. At step $t$, the state $s_t$ consists of the current node $c_t=\pi_t$, the start node $s_{\text{start}}=\pi_1$, and the set of unvisited nodes $U_t$. The true optimal action‑value for a candidate $j\in U_t$, under cost minimization, follows the Bellman equation:
\begin{equation}
Q^*(j \mid s_t) = D_{c_t,j} + V^*(j,\,U_t\setminus\{j\}), \qquad V^*(i,\emptyset) = D_{i,s_{\text{start}}}. \label{eq:bellman}
\end{equation}

Equation~\eqref{eq:bellman} shows that the quality of choosing $j$ is the sum of the immediate directed edge cost $D_{c_t,j}$ and the optimal cost‑to‑go from $j$, where the terminal condition explicitly involves the return‑to‑start cost. Therefore, a well‑inducted decoder should make the final score depend on (i) the immediate transition $c_t\to j$, (ii) the return boundary $j\to s_{\text{start}}$, and (iii) a suitable surrogate for the continuation value $V^*$. The rest of this section shows that standard neural decoders often fail to provide the first ingredient in a scale‑robust manner, and how a simple, explicit edge‑aware bias resolves this issue.

\subsection{Baseline Decoder and the Representation–Decision Mismatch}
\label{subsec:radar-baseline}

We adopt the leading SVD/Sinkhorn asymmetric backbone from RADAR~\cite{yi2026radar} as a controlled testbed (see Appendix~\ref{app:radar-baseline} for full equations). Given the normalized cost matrix $\hat{D}$ (instance‑level z‑score), the encoder produces node embeddings $\{h_i\}_{i=1}^n$ through truncated SVD initialization and asymmetry‑aware attention. At step $t$, the decoder forms a context vector $z_t$ from the partial tour and the embeddings, and scores a feasible candidate $j$ via the dot‑product compatibility
\begin{equation}
s_t^{\mathrm{base}}(j) = \frac{z_t^{\top} k_j}{\sqrt{d}}, \label{eq:base}
\end{equation}
where $k_j$ is a linear projection of $h_j$. Although $z_t$ conditions on $c_t$ and $k_j$ comes from an asymmetry‑aware encoder, the score \eqref{eq:base} is fundamentally a \emph{context–node compatibility} term; the directed edge $c_t\to j$ is not an explicit argument of the function.

We can decompose the baseline score into interpretable components. Because both $z_t$ and $k_j$ are computed from the same instance matrix $\hat{D}$, the dot product implicitly captures some portion of the edge cost. Formally, write
\begin{equation}
s_t^{\mathrm{base}}(j) = \underbrace{g(j; s_t)}_{\text{state–node interaction}} \;+\; \underbrace{\widehat D_{c_t,j}^{\mathrm{\,(emb)}}}_{\mathclap{\text{implicit estimate of }\hat{D}_{c_t,j}}} \;+\; \eta_j, \label{eq:decomp}
\end{equation}
where $g$ collects all non‑edge information (e.g., node centrality, depot bias) and $\eta_j$ is irreducible embedding noise. The implicit edge estimate $\widehat D_{c_t,j}^{\mathrm{\,(emb)}}$ is not a simple linear readout; it emerges from the high‑dimensional distributed representations of the encoder. Crucially, the encoder is trained exclusively on instances of a particular size, e.g., $n_{\text{train}}\!=\!100$. When applied to larger instances, the embedding space undergoes a distribution shift. This causes the implicit estimate to deviate from the true normalized cost:
\begin{equation}
\widehat D_{c_t,j}^{\mathrm{\,(emb)}} = \hat{D}_{c_t,j} + \varepsilon_j(n), \qquad
\mathbb{E}\bigl[|\varepsilon_j(n)|\bigr] \uparrow \text{ as } n \text{ increases}. \label{eq:error}
\end{equation}
The error $\varepsilon_j(n)$ arises because the encoder's representation entangles the edge cost with size‑dependent patterns (e.g., average degree, attention spread) that are not invariant across scales.
Consequently, on large instances the baseline decoder's ability to correctly order candidates by their true transition cost degrades. This is the \emph{representation–decision mismatch}: directed edge information is encoded upstream but fails to reliably reach the final decision score under scale shift.

\begin{figure}[t]
    \centering
    \includegraphics[width=\textwidth]{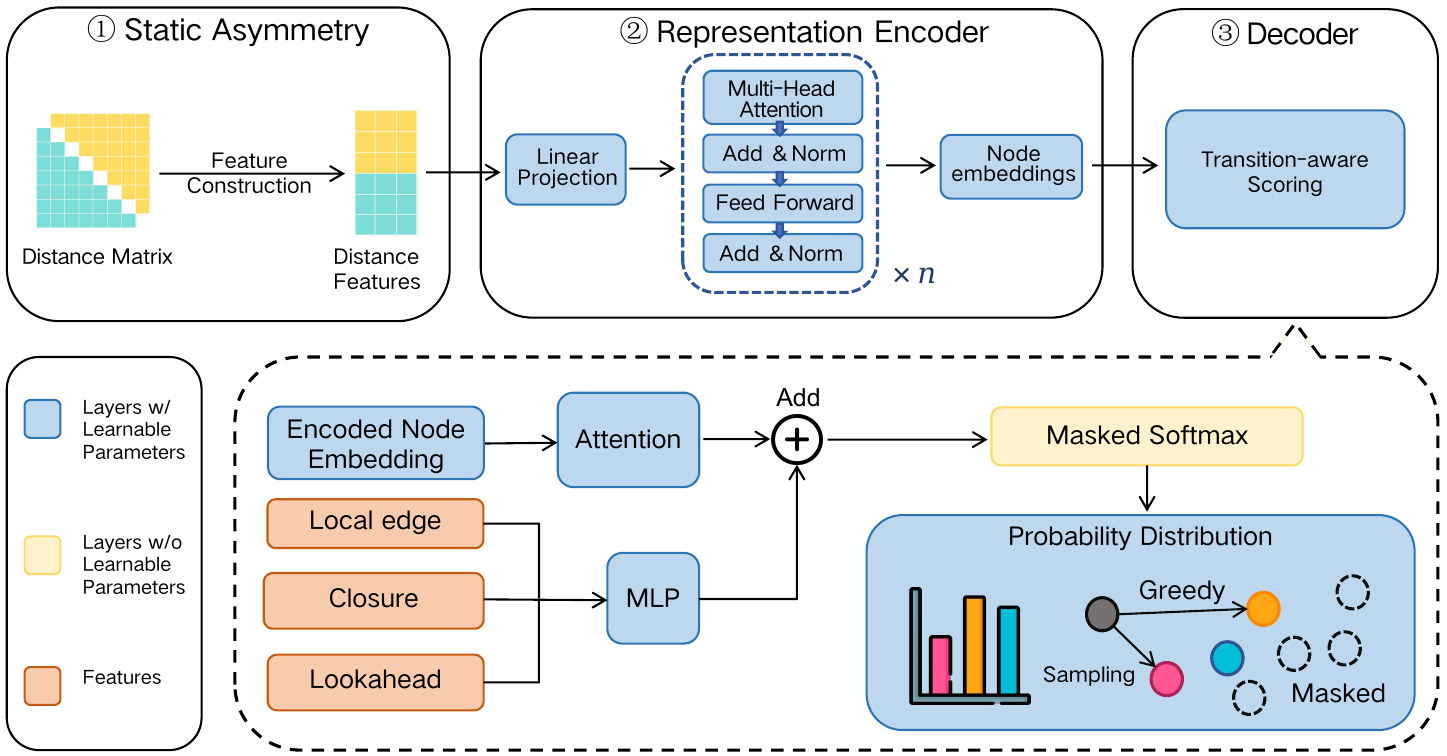}
    \caption{ATSP framework for decoder-side edge-aware scoring. The controlled
    SVD/Sinkhorn asymmetric representation pipeline encodes directed matrix
    information into node embeddings, while the proposed decoder augments the
    final score with local edge, closure, and lookahead descriptors before
    masked action selection.}
    \label{fig:framework-atsp}
\end{figure}

\subsection{Edge‑Aware Decoder Bias: Explicit Transition Features}
\label{subsec:edge-aware-decoder}

To correct this mismatch, we introduce an \emph{edge‑aware bias} that directly exposes transition‑specific quantities in the final score. Let $\phi_t(j)$ be a transition descriptor computed deterministically from $\hat{D}$, the current node $c_t$, the candidate $j$, and the start $s_{\text{start}}$. We design $\phi_t(j)$ to mirror the Bellman decomposition~\eqref{eq:bellman}:
\begin{align}
\phi_t(j) &= \bigl[\, \underbrace{\hat{D}_{c_t,j},\; \hat{D}_{j,c_t},\; \hat{D}_{c_t,j}-\hat{D}_{j,c_t}}_{\text{local edge } \phi_t^{\mathrm{loc}}}\, \| \,
\underbrace{\hat{D}_{j,s_{\text{start}}}}_{\text{closure } \phi_t^{\mathrm{close}}}\, \| \,
\underbrace{\min_{\ell\neq j}\hat{D}_{j\ell},\; \min_{\ell\neq j}\hat{D}_{\ell j}}_{\text{static lookahead } \phi_t^{\mathrm{look}}}\, \bigr]. \label{eq:phi}
\end{align}
The local edge terms provide the immediate directed cost and a reverse‑edge contrast, enabling the score to directly penalize expensive transitions. The closure term supplies the terminal return‑to‑start boundary condition, and the static lookahead terms give coarse, scale‑invariant estimates of connectivity after visiting $j$ (cheap to precompute as they do not depend on $U_t$).

The bias is produced by a small multi‑layer perceptron:
\begin{equation}
b_t(j) = \mathrm{MLP}_{\mathrm{edge}}\bigl(\phi_t(j)\bigr), \label{eq:bias}
\end{equation}
and added to the baseline score to yield the final logit
\begin{equation}
\tilde{s}_t(j) = s_t^{\mathrm{base}}(j) + b_t(j). \label{eq:aug}
\end{equation}
The policy is obtained by masking infeasible nodes, clipping, and softmax as usual.
\begin{equation}
p_t(j) = \mathrm{Softmax}\Bigl(C\tanh\bigl(\tilde{s}_t(j)\bigr) + \mathrm{mask}_t(j)\Bigr).
\end{equation}

Figure~\ref{fig:framework-atsp} illustrates how the edge‑aware bias integrates into the full autoregressive pipeline. The bias module is trained end‑to‑end together with the encoder and the base decoder, with no auxiliary supervision beyond the reinforcement‑learning objective.

\subsection{Scale-Robust Regularization via Edge-Cost Decoupling}

The effectiveness of the edge‑aware bias under scale shift stems from it providing a \emph{noise‑free, scale‑invariant} pathway for the critical directed edge signal. Even a minimal linear configuration of the MLP can learn
\begin{equation}
b_t(j) \approx -\alpha\,\hat{D}_{c_t,j} + \beta^{\top} \tilde\phi_t(j) + \delta, \quad \alpha>0, \label{eq:linear_bias}
\end{equation}
where $\tilde\phi_t(j)$ contains the remaining features. The negative sign follows from the cost‑minimization objective. Substituting~\eqref{eq:decomp} and~\eqref{eq:linear_bias} into~\eqref{eq:aug} gives
\begin{align}
\tilde{s}_t(j) = g(j;s_t) + \bigl(\hat{D}_{c_t,j} + \varepsilon_j(n)\bigr) + \bigl(-\alpha\hat{D}_{c_t,j} + \beta^{\top}\tilde\phi_t(j) + \delta\bigr) + \eta_j. \label{eq:combined}
\end{align}
The explicit $-\alpha\hat{D}_{c_t,j}$ term directly counterweights the implicit estimate, reducing the overall dependence on the embedding error $\varepsilon_j(n)$. For a pair of candidates $j_1,j_2$, the score difference becomes
\begin{align}
\Delta\tilde{s} = \Delta g + (1-\alpha)(\hat{D}_{c_t,j_1}-\hat{D}_{c_t,j_2}) + (\varepsilon_{j_1}(n)-\varepsilon_{j_2}(n)) + \beta^{\top}\Delta\tilde\phi + \Delta\eta. \label{eq:pairwise}
\end{align}
When $\alpha$ is learned sufficiently large, the term $(1-\alpha)(\hat{D}_{c_t,j_1}-\hat{D}_{c_t,j_2})$ can dominate, making the relative ranking robust to the embedding noise $\varepsilon$. Moreover, the closure and lookahead features $\tilde\phi$ are themselves deterministic functions of the cost matrix, they provide a simple and direct pathway for edge and boundary information that would otherwise have to be inferred from embeddings.

From a learning‑theoretic perspective, the bias module acts on a fixed‑dimensional feature vector whose relationship to the optimal action is nearly monotonic and independent of $n$, ensuring that the learned mapping provides an inductive bias across scales. In contrast, the baseline decoder must recover the same edge signal from high‑dimensional embeddings, a task whose difficulty increases with the distribution shift. The explicit bias therefore acts as a \emph{scale‑robust regularizer} that decouples the decision‑critical edge cost from the learned representation space.

\subsection{Complexity and Implementation}

Evaluating the bias for all feasible candidates adds $O(Bn)$ work per decoding step for a batch of size $B$, dominated by the MLP forward pass. Over a full rollout, this yields an $O(B n^2)$ overhead, which, while noticeable, is manageable (e.g., 43 vs.\ 11 minutes on ATSP‑1000 in our experiments). We cache static quantities (transpose matrices, static minima) to minimize repeated computation. Further speed‑optimization is left for future work.

\noindent\textbf{Extension to ACVRP.} For the ACVRP, the closure term is replaced by the candidate‑to‑depot cost $\hat{D}_{j,0}$. The same score‑level augmentation is applied with this minimal adaptation (see Appendix~\ref{app:acvrp-extension}).

\section{Experiments}
\label{sec:experiments}

\subsection{Setup}

\paragraph{Scale-generalization protocol.}
Following the scale-generalization protocol of RADAR~\citep{yi2026radar}, we train all newly evaluated decoder variants from scratch on 100-node synthetic instances and evaluate them zero-shot on target sizes \(n \in \{100,200,500,1000\}\). Training instances are generated online, with 10,000 sampled episodes per epoch for 2,100 epochs. For each target size, evaluation uses the corresponding fixed held-out test set of 1,000 instances. Unless otherwise stated, test-time augmentation is disabled, and each reported value is averaged over the full test set. For both ATSP and ACVRP, we report the average objective value (\textbf{Obj}), the percentage gap relative to the LKH3 reference objective (\textbf{Gap}), and total evaluation time (\textbf{Time}; units shown in each table). The table headers indicate the preferred direction for each metric. Architecture, optimizer, edge-feature, and compute details are provided in Appendix~\ref{app:reproducibility-details} and Table~\ref{tab:app-hyperparams}.

\subsection{Baselines}

The external comparisons use the baseline set reported by RADAR~\citep{yi2026radar} as benchmark context under the corresponding benchmark setting. For ATSP, this set includes LKH3 variants~\citep{helsgaun2017lkh3}, GLOP~\citep{ye2024glop}, UDC-x250~\citep{zheng2024udc}, MatNet-family baselines~\citep{kwon2021matnet}, MatPOENet-8x~\citep{pan2025unico}, ICAM~\citep{zhou2024icam}, ELG~\citep{gao2024elg}, ReLD~\citep{huang2025reld}, and the RADAR reference~\citep{yi2026radar}. For ACVRP, the reference set follows the corresponding RADAR table, including HGS, LKH3, MatNet-family, ReLD, and RADAR reference entries.

The controlled comparison keeps the SVD/Sinkhorn asymmetric backbone fixed and modifies only the final decoder logits through decoder-side edge-aware scoring. This design isolates the contribution of decision-time transition features from representation-side controls. Table~\ref{tab:main-results} compares the proposed decoder with the RADAR reference entries, while Table~\ref{tab:ablation-results} varies the decoder edge-feature set within the same fixed-backbone family.

\subsection{Results}

\subsubsection{ATSP and ACVRP Results}

\begin{figure}[t]
    \centering
    \includegraphics[width=0.8\textwidth]{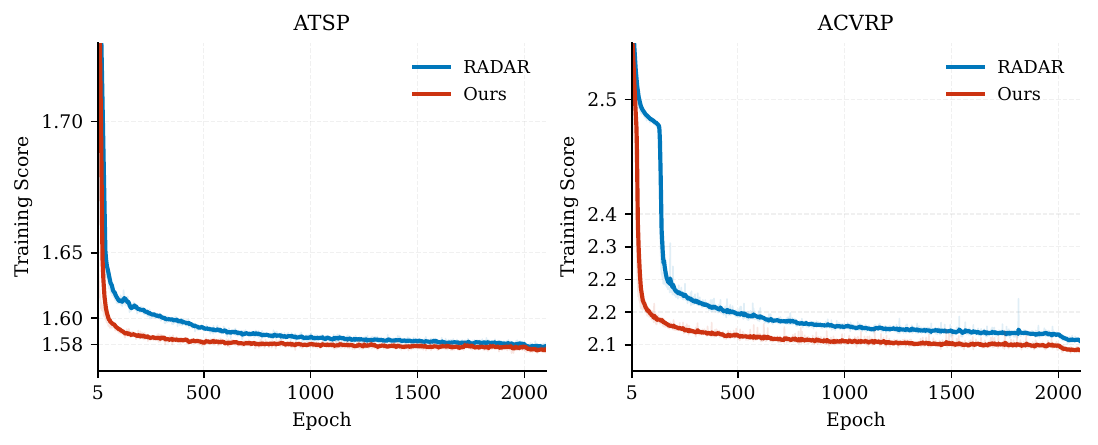}
    \caption{Training score curves for RADAR and our method on ATSP and ACVRP. The left panel reports ATSP and the right panel reports ACVRP; faint traces show raw epoch scores and solid curves show exponential moving average scores.}
    \label{fig:training-score}
\end{figure}

Figure~\ref{fig:training-score} reports training-score dynamics for RADAR and our method on ATSP and ACVRP. The curves provide an optimization-level view of the same decoder intervention: the representation backbone is fixed, and only decoder-side edge-aware scoring changes the final decoder logits. The solid EMA traces show that our method improves the training trajectory on both tasks, while the raw traces indicate epoch-level variability.

Table~\ref{tab:main-results} reports objective, gap, and runtime for the corresponding held-out benchmark evaluation. The upper block reports ATSP results, and the lower block reports ACVRP results under the corresponding benchmark setting. The decoder ablation below varies only the decoder edge-feature set within the same fixed-backbone family.

\begin{table*}[t]
    \centering
    \scriptsize
    \caption{ATSP and ACVRP benchmark comparison across 100/200/500/1000-node instances. Baseline rows and annotations are reproduced from RADAR~\citep{yi2026radar}; Time uses the units shown in each entry. Arrows indicate preferred direction.}
    \label{tab:main-results}
    \label{tab:acvrp-results}
    \resizebox{\textwidth}{!}{%
    \begin{tabular}{l|ccc|ccc|ccc|ccc}
        \toprule
        & \multicolumn{3}{c|}{\textbf{Testing} (1k instances)}
        & \multicolumn{9}{c}{\textbf{Generalization} (1k instances)} \\
        & \multicolumn{3}{c|}{ATSP100}
        & \multicolumn{3}{c|}{ATSP200}
        & \multicolumn{3}{c|}{ATSP500}
        & \multicolumn{3}{c}{ATSP1000} \\
        Method
        & Obj.$\downarrow$ & Gap$\downarrow$ & Time$\downarrow$
        & Obj.$\downarrow$ & Gap$\downarrow$ & Time$\downarrow$
        & Obj.$\downarrow$ & Gap$\downarrow$ & Time$\downarrow$
        & Obj.$\downarrow$ & Gap$\downarrow$ & Time$\downarrow$ \\
        \midrule
        LKH-100
        & 1.5643 & 0.00\% & 1.08m
        & 1.5721 & 0.00\% & 2.44m
        & 1.5763 & 0.00\% & 7.32m
        & 1.5739 & 0.00\% & 21.92m \\
        LKH-1000
        & 1.5643 & * & 5.91m
        & 1.5721 & * & 16.40m
        & 1.5763 & * & 44.45m
        & 1.5739 & * & 1.71h \\
        \midrule
        GLOP$^{+}$
        & 1.8848 & 20.49\% & 2.75m
        & 2.0584 & 30.93\% & 3.07m
        & 2.2345 & 41.76\% & 3.72m
        & 2.3429 & 48.86\% & 7.29m \\
        UDC-$x$250$^{+}$ ($\alpha=50$)
        & 1.5921 & 1.78\% & 1.02h
        & 1.7577 & 11.81\% & 2.04h
        & 2.3810 & 51.05\% & 6.41h
        & 3.0734 & 95.27\% & 19.12h \\
        \midrule
        Matnet
        & 1.6473 & 5.31\% & 0.03m
        & 3.1925 & 103.07\% & 0.14m
        & -- & -- & --
        & -- & -- & -- \\
        Matnet$^{+}$
        & 1.6161 & 3.32\% & 0.03m
        & 1.9111 & 21.56\% & 0.14m
        & -- & -- & --
        & -- & -- & -- \\
        Matnet-Single (One-hot)$^{+}$
        & 1.5995 & 2.25\% & 0.02m
        & 1.6727 & 6.40\% & 0.12m
        & -- & -- & --
        & -- & -- & -- \\
        Matnet-Single (Random)$^{+}$
        & 1.5969 & 2.08\% & 0.02m
        & 1.6543 & 5.23\% & 0.13m
        & 1.8610 & 18.10\% & 1.34m
        & 2.1821 & 38.64\% & 11.11m \\
        MatPOENet-8x$^\star$
        & 1.8719 & 19.67\% & 0.54m
        & 2.7033 & 71.95\% & 2.33m
        & 4.1189 & 161.30\% & 21.87m
        & 5.5072 & 249.91\% & 2.98h \\
        ICAM$^{+}$
        & 1.6580 & 5.99\% & 0.01m
        & 1.8471 & 17.49\% & 0.06m
        & 2.4592 & 56.01\% & 0.73m
        & 2.9069 & 84.69\% & 9.80m \\
        ELG$^{+}$
        & 1.5982 & 2.17\% & 0.06m
        & 1.6423 & 4.47\% & 0.27m
        & 1.7456 & 10.74\% & 2.60m
        & 1.8441 & 17.17\% & 22.31m \\
        ReLD$^{+}$
        & 1.5900 & 1.64\% & 0.03m
        & 1.6310 & 3.75\% & 0.15m
        & 1.7873 & 13.39\% & 1.49m
        & 2.0723 & 31.67\% & 12.18m \\
        RADAR$^{+}$
        & 1.5756 & 0.72\% & 0.04m
        & 1.5879 & 1.01\% & 0.15m
        & 1.6098 & 2.13\% & 1.45m
        & 1.6389 & 4.13\% & 11.57m \\
        \midrule
        Ours
        & \textbf{1.5741} & \textbf{0.63\%} & 0.05m
        & \textbf{1.5852} & \textbf{0.83\%} & 0.38m
        & \textbf{1.6032} & \textbf{1.71\%} & 5.31m
        & \textbf{1.6170} & \textbf{2.73\%} & 43.44m \\
        \bottomrule
    \end{tabular}
    }
    \vspace{0.35em}
    \resizebox{\textwidth}{!}{%
    \begin{tabular}{l|ccc|ccc|ccc|ccc}
        \toprule
        & \multicolumn{3}{c|}{\textbf{Testing} (1k instances)}
        & \multicolumn{9}{c}{\textbf{Generalization} (1k instances)} \\
        & \multicolumn{3}{c|}{ACVRP100}
        & \multicolumn{3}{c|}{ACVRP200}
        & \multicolumn{3}{c|}{ACVRP500}
        & \multicolumn{3}{c}{ACVRP1000} \\
        Method
        & Obj.$\downarrow$ & Gap$\downarrow$ & Time$\downarrow$
        & Obj.$\downarrow$ & Gap$\downarrow$ & Time$\downarrow$
        & Obj.$\downarrow$ & Gap$\downarrow$ & Time$\downarrow$
        & Obj.$\downarrow$ & Gap$\downarrow$ & Time$\downarrow$ \\
        \midrule
        LKH-100
        & 2.2526 & 6.05\% & 2.20m
        & 2.2245 & 2.77\% & 3.24m
        & 2.4155 & 3.20\% & 11.58m
        & 2.5092 & 2.84\% & 1.00h \\
        LKH-1000
        & 2.1635 & 1.86\% & 17.64m
        & 2.1807 & 0.75\% & 25.32m
        & 2.3605 & 0.86\% & 1.06h
        & 2.4569 & 0.69\% & 2.86h \\
        LKH-10000
        & 2.1240 & 0.00\% & 2.79h
        & 2.1645 & 0.00\% & 4.25h
        & 2.3405 & 0.00\% & 10.29h
        & 2.4400 & 0.00\% & 36.25h \\
        \midrule
        HGS-Short$^{\#}$
        & 2.1614 & 1.76\% & 16.67m
        & 2.0806 & -3.88\% & 25.38m
        & 2.3011 & -1.68\% & 1.08h
        & 2.1094 & -13.55\% & 2.99h \\
        HGS-Long$^{\#}$
        & 2.0942 & -1.40\% & 2.78h
        & 1.9733 & -8.83\% & 4.24h
        & 2.1451 & -8.35\% & 10.47h
        & 1.9792 & -18.89\% & 36.33h \\
        \midrule
        Matnet (Demand)$^{+}$
        & 2.1968 & 3.42\% & 0.04m
        & 3.1620 & 46.10\% & 0.20m
        & 4.4847 & 91.58\% & 1.92m
        & 5.7523 & 135.76\% & 13.59m \\
        Matnet-Single (Demand)$^{+}$
        & 2.1821 & 2.73\% & 0.03m
        & 2.6283 & 21.43\% & 0.15m
        & 3.4994 & 49.46\% & 1.67m
        & 4.3736 & 79.25\% & 11.79m \\
        Matnet-Single (Random)$^{+}$
        & 2.1813 & 2.70\% & 0.03m
        & 2.2372 & 3.36\% & 0.15m
        & 3.3697 & 43.98\% & 1.73m
        & 3.9904 & 63.52\% & 11.38m \\
        ReLD$^{+}$
        & 2.1656 & 1.96\% & 0.04m
        & 2.1635 & -0.05\% & 0.18m
        & 3.4507 & 47.40\% & 1.96m
        & 4.7424 & 94.45\% & 13.80m \\
        RADAR$^{+}$
        & 2.1588 & 1.64\% & 0.05m
        & 2.1483 & -0.75\% & 0.14m
        & 2.4198 & 3.39\% & 1.75m
        & 2.4634 & 0.96\% & 11.60m \\
        \midrule
        Ours
        & \textbf{2.1444} & \textbf{0.96\%} & 0.09m
        & \textbf{2.1064} & \textbf{-2.68\%} & 0.33m
        & \textbf{2.3944} & \textbf{2.30\%} & 4.79m
        & \textbf{2.1148} & \textbf{-13.33\%} & 36.30m \\
        \bottomrule
    \end{tabular}
    }
    \begin{flushleft}
    \scriptsize
    \textbf{Note.} $^{\star}$ denotes evaluation using the authors' official checkpoints; $^{+}$ indicates training and testing with z\mbox{-}score normalization. $^{\#}$ indicates HGS rows reported by RADAR~\citep{yi2026radar} with infeasible ACVRP solutions under the corresponding budgets; we therefore treat them only as external references and do not use them for gap computation. Results for UDC and ICAM differ from their original reports because their mixed\mbox{-}size training scheme is disabled to align with the single-size training condition used in this benchmark. MatNet with one\mbox{-}hot node embeddings fails to generalize to $N\in\{500,1000\}$, as the identity embedding is tied to a fixed 256\mbox{-}dimensional space. The boldface indicates the best result among learning-based methods.
    \end{flushleft}
\end{table*}

\paragraph{ATSP.}
In ATSP, each construction step corresponds to selecting a directed transition. The upper block of Table~\ref{tab:main-results} evaluates decoder-side edge-aware scoring while keeping the SVD-backed backbone fixed. The gain is modest near the training scale and larger under zero-shot scaling. Relative to the RADAR reference, our method reduces the ATSP-100 gap from $0.72\%$ to $0.63\%$. At ATSP-500 and ATSP-1000, the gap decreases from $2.13\%$ to $1.71\%$ and from $4.13\%$ to $2.73\%$, respectively. This pattern is consistent with the claim that decision-time transition features are under-exposed in the final decoder score. The cost is higher inference time: at ATSP-1000, evaluation time increases from $11.57$ to $43.44$ minutes.

\paragraph{ACVRP.}
ACVRP evaluates the same score-level modification under a richer routing state, where the decoder must account for asymmetric travel cost together with depot return, demand feasibility, and route continuation. The lower block of Table~\ref{tab:main-results} uses the corresponding RADAR benchmark setting as external reference and keeps the proposed variant within the same fixed-backbone family. Relative to the RADAR reference, our method improves the average objective at all evaluated sizes: from $2.1588$ to $2.1444$ on ACVRP-100, from $2.1483$ to $2.1064$ on ACVRP-200, from $2.4198$ to $2.3944$ on ACVRP-500, and from $2.4634$ to $2.1148$ on ACVRP-1000. These results show that the final-score modification remains useful when additional routing constraints enter the decoder state.

\subsubsection{ATSP Decoder Ablation}
\label{sec:atsp-ablation}

Table~\ref{tab:ablation-results} evaluates the decoder features motivated by the
Bellman-style cost-to-go view in Section~\ref{subsec:edge-aware-decoder}. The
local edge terms expose the current directed cost and reverse-edge contrast, the
closure term provides a candidate-specific return-to-start boundary cue, and the
static lookahead terms provide lightweight one-edge proxies for continuation
after selecting \(j\). All rows keep the SVD/Sinkhorn backbone fixed and vary
only these final-score features, so the comparison isolates which decision-time
quantities are needed by the decoder.

\begin{table*}[t]
    \centering
    \footnotesize
    \caption{ATSP decoder-edge feature ablation on the fixed SVD/Sinkhorn backbone. Each row toggles the local edge, closure, and lookahead terms defined in Section~\ref{subsec:edge-aware-decoder}; Time is reported in minutes. Arrows indicate preferred direction.}
    \label{tab:ablation-results}
    \resizebox{\textwidth}{!}{%
    \begin{tabular}{lccc*{4}{ccc}}
        \toprule
        & & & & \multicolumn{3}{c}{ATSP-100} & \multicolumn{3}{c}{ATSP-200} & \multicolumn{3}{c}{ATSP-500} & \multicolumn{3}{c}{ATSP-1000} \\
        \cmidrule(lr){5-7} \cmidrule(lr){8-10} \cmidrule(lr){11-13} \cmidrule(lr){14-16}
        Method & Local edge & Closure & Lookahead & Obj$\downarrow$ & Time$\downarrow$ & Gap$\downarrow$ & Obj$\downarrow$ & Time$\downarrow$ & Gap$\downarrow$ & Obj$\downarrow$ & Time$\downarrow$ & Gap$\downarrow$ & Obj$\downarrow$ & Time$\downarrow$ & Gap$\downarrow$ \\
        \midrule
        RADAR & $\times$ & $\times$ & $\times$ & 1.5756 & 0.04 & 0.72\% & 1.5879 & 0.15 & 1.01\% & 1.6098 & 1.45 & 2.13\% & 1.6389 & 11.57 & 4.13\% \\
        Ours & $\checkmark$ & $\checkmark$ & $\checkmark$ & 1.5741 & 0.05 & 0.63\% & \textbf{1.5852} & 0.38 & \textbf{0.83\%} & \textbf{1.6032} & 5.31 & \textbf{1.71\%} & \textbf{1.6170} & 43.44 & \textbf{2.73\%} \\
        Ours & $\times$ & $\checkmark$ & $\checkmark$ & 1.5766 & 0.05 & 0.79\% & 1.5909 & 0.36 & 1.20\% & 1.6188 & 4.76 & 2.70\% & 1.6533 & 40.67 & 5.04\% \\
        Ours & $\checkmark$ & $\times$ & $\checkmark$ & \textbf{1.5730} & 0.07 & \textbf{0.56\%} & 1.5857 & 0.36 & 0.87\% & 1.6101 & 4.63 & 2.14\% & 1.6340 & 40.40 & 3.82\% \\
        Ours & $\checkmark$ & $\checkmark$ & $\times$ & 1.5739 & 0.05 & 0.61\% & 1.5855 & 0.37 & 0.85\% & 1.6057 & 4.71 & 1.87\% & 1.6217 & 41.50 & 3.03\% \\
        \bottomrule
    \end{tabular}
    }
\end{table*}

The full decoder improves over RADAR at every evaluated size, with the largest
gap reduction at ATSP-1000, from $4.13\%$ to $2.73\%$. The strongest negative
result is the row without local edge terms: even with closure and lookahead
retained, the ATSP-1000 gap increases to $5.04\%$, worse than the RADAR
reference.
This pattern is consistent with the cost-to-go relation
\(C^*(i,j,U)=D_{ij}+V^*(j,U\setminus\{j\})\). If the immediate directed cost
\(D_{ij}\) is absent from the final decoder score, the remaining boundary and
continuation proxies do not reliably recover the correct transition preference
from the SVD-backed representation alone.

Closure and lookahead play different roles once the local transition term is
present. Removing closure gives the best ATSP-100 gap in the table, but the
advantage disappears under scale generalization: the ATSP-1000 gap rises from
$2.73\%$ to $3.82\%$. This is consistent with the Bellman-style boundary
condition \(V^*(i,\varnothing)=D_{is}\), where the candidate-to-start cost
becomes a direct terminal quantity rather than a generic node feature. Removing
lookahead causes a smaller but consistent large-scale degradation, raising the
ATSP-500 and ATSP-1000 gaps from $1.71\%$ to $1.87\%$ and from $2.73\%$ to
$3.03\%$, respectively. The static lookahead terms therefore act as heuristic
continuation cues rather than exact estimates of
\(V^*(j,U\setminus\{j\})\).

Taken together, the ablation supports a transition cost-to-go interpretation of
the decoder. The local edge term supplies the immediate directed cost, closure
supplies a boundary-condition scalar, and lookahead supplies coarse continuation
information. The subsequent diagnostics separate this transfer-level evidence
from the narrower question of whether the final decoder score responds correctly
to perturbations of the current directed edge.

We present supplementary comparisons in the appendices: static versus masked lookahead variants appear in Appendix~\ref{sec:lookahead-mask-comparison}, where masked lookahead yields a marginal ATSP-100 improvement at the cost of substantially increased runtime, while underperforming at larger scales, confirming static lookahead as the preferred continuation cue. Appendix~\ref{app:pomo-tsp} and Appendix~\ref{app:elg-decoder-bias} provide additional experimental results on POMO-style symmetric TSP and ELG-style large-scale routing benchmarks, respectively.

\subsection{Directed-transition Diagnostics}
\label{sec:transition-diagnostics}

Benchmark ablations identify useful decoder features but do not directly test whether the final logit depends on the immediate directed transition. The node-centric form of the baseline decoder—scoring candidate \(j\) from context and node embedding while the action is the transition \(i \to j\)—suggests that asymmetric edge information may be encoded yet weakly expressed in the final score.

We design a four-part diagnostic suite on actual ATSP decoding states. For each state, we record the start node, current node \(i=c_t\), visited set, and feasible candidates. \textit{Current SignAcc} increases \(D_{ij}\) and checks whether the logit of candidate \(j\) decreases. \textit{Reverse SignAcc} increases \(D_{ji}\) and serves as a signed-response control, since \(D_{ji}\) is not the cost of action \(i\to j\). \textit{Antisym SignAcc} increases \(D_{ij}\) while decreasing \(D_{ji}\), testing whether the final score reacts to directed asymmetry in the current transition. \textit{Local PairAcc} does not perturb the matrix; it samples matched candidate pairs with similar closure and lookahead quantities and measures whether the candidate with the smaller current directed edge receives the higher logit.

For perturbation‑based metrics, each counterfactual matrix is fed through the full encoder–decoder pipeline, allowing RADAR to respond via its representation. Appendix~\ref{app:transition-diagnostics} details the state collection protocol and an additional swap control.

\begin{figure*}[t]
    \centering
    \includegraphics[width=\textwidth]{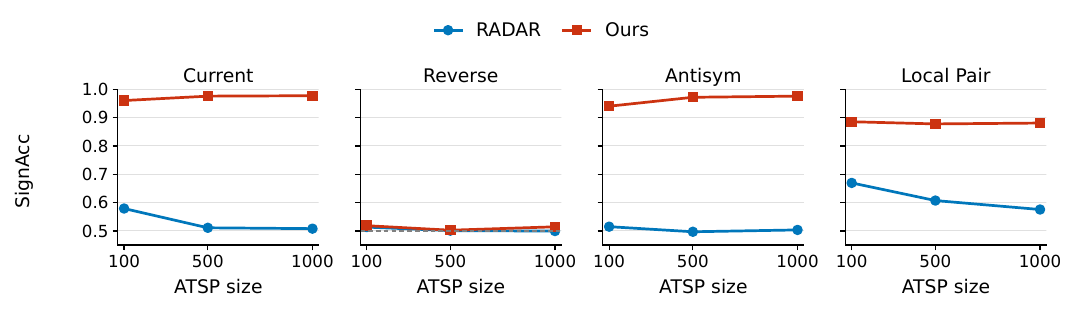}
    \caption{Directed‑transition diagnostics across ATSP sizes. Our method exhibits stronger response to current directed‑edge and antisymmetric perturbations than RADAR, while Reverse SignAcc remains near the 0.5 signed‑response control level. Our method also improves Local PairAcc at all tested scales.}
    \label{fig:transition-diagnostics}
\end{figure*}

Figure~\ref{fig:transition-diagnostics} shows that the improvement is concentrated on decision‑relevant directed transitions: our method responds strongly to perturbations of the current edge and to antisymmetric edge changes, while remaining near chance on the reverse‑edge control.

These diagnostics support a narrow mechanism claim. RADAR is not edge‑blind end‑to‑end, but its final score exhibits weaker and less stable sensitivity to the current directed transition, especially at larger scales. Edge‑aware decoding directly strengthens this decision‑time transition sensitivity. These tests mainly verify sensitivity to the immediate directed edge.

\section{Conclusion}
\label{sec:conclusion}

This paper identifies representation–decision mismatch as a bottleneck in neural asymmetric routing: directed cost information is encoded upstream, yet the final decoder score remains largely parameterized as context–node compatibility. We show that exposing transition-specific information directly in the decoder score mitigates this mismatch.

Empirically, decoder-side edge-aware scoring improves over the RADAR reference on ATSP across 100–1000 nodes, with larger gains under scale generalization. On ACVRP, the same score-level modification improves performance under a richer routing state involving depot return, demand feasibility, and route continuation. Ablations confirm that the local directed edge is the most critical component, while closure and static lookahead act as useful continuation cues. Directed-transition diagnostics provide targeted evidence that edge-aware scores respond more reliably to changes in the current directed edge.

Beyond these results, we advocate a decoder-design principle: constructive solvers should not rely solely on node compatibility but also expose action-level quantities reflecting immediate cost and lightweight proxies for future cost-to-go. More broadly, this suggests a route for designing decoder augmentations by identifying transition-level cues implied by the problem’s cost-to-go structure.

\paragraph{Limitations}
The proposed decoder bias adds substantial inference overhead at large scale, making efficiency an important practical trade-off. Closure and lookahead remain heuristic proxies rather than fully isolated mechanisms.

\paragraph{Broader impact}
Better routing heuristics can improve solution quality and operational efficiency, but our results are limited to synthetic benchmarks and should not be interpreted as deployment-ready guarantees for real logistics systems. Practical use would require stronger robustness validation and domain-specific safety checks.

\bibliographystyle{plainnat}
\bibliography{references}

\newpage

\appendix

\section{Reproducibility Details}
\label{app:reproducibility-details}

The experiments are implemented as patches to the RADAR codebase \citep{yi2026radar}. The repository layout used by the experiments has separate task directories \path{atsp/} and \path{acvrp/}, each containing the corresponding environment, model, trainer, tester, \path{train.py}, and \path{test.py}. The decoder-aware changes are localized to the task models and trainers, with the shared policy-loss helper in \path{utils/rl_objectives.py}. The preset scripts record the resolved model and trainer parameters in each result directory and save checkpoints every 100 epochs; all newly trained rows use checkpoint 2100.

\paragraph{Data.}
Training instances are generated online at size 100. Table~\ref{tab:app-data-artifacts} summarizes the fixed evaluation files and upstream RADAR artifacts used for testing and initialization. ATSP costs are divided by $10^6$ when loaded, and both ATSP and ACVRP models use instance-level z-score normalization for the directed distance matrix before constructing SVD and edge features. The RADAR OpenReview paper is CC BY 4.0; separate Google Drive artifact and upstream codebase licenses were not verified.

\paragraph{Hyperparameters.}
Table~\ref{tab:app-hyperparams} summarizes the default configuration shared across ATSP and ACVRP unless explicitly ablated. The full edge descriptor used by the decoder is
\[
\left[
\hat D_{ij},\hat D_{ji},\hat D_{ij}-\hat D_{ji},
\hat D_{js},
\min_{\ell\ne j}\hat D_{j\ell},
\min_{\ell\ne j}\hat D_{\ell j}
\right]
\]
for ATSP, where $s$ is the start node.
For ACVRP, node 0 is the depot and the closure term is replaced by $\hat D_{j0}$.
In the ATSP ablations, the \texttt{no-lookahead} variant removes the two minimum-edge terms from this descriptor.

\paragraph{Compute and statistical scope.}
The main training runs were executed on a single NVIDIA RTX 5090 GPU. The main ATSP and ACVRP Ours runs took 24.86 and 28.08 GPU-hours, respectively. Evaluation-time units are indicated in the corresponding tables. The newly trained rows use the checkpoint and fixed 1,000-instance test files specified above.

\section{ACVRP Extension}
\label{app:acvrp-extension}

The ACVRP experiments in the main text use the same score-level idea with minimal task-specific adaptation. The encoder remains matrix-based and SVD-initialized, but the decoder query is conditioned on both the current node representation and the remaining vehicle load. At decision step $t$, if $i_t$ is the current node and $0$ denotes the depot, the ACVRP edge-aware descriptor becomes
\begin{equation}
\phi_t^{\mathrm{acvrp}}(j)=
\left[
\hat D_{i_t j},
\hat D_{j i_t},
\hat D_{i_t j}-\hat D_{j i_t},
\hat D_{j0},
\min_{\ell\ne j}\hat D_{j\ell},
\min_{\ell\ne j}\hat D_{\ell j}
\right].
\end{equation}

\section{RADAR Baseline}
\label{app:radar-baseline}

This appendix records the SVD/Sinkhorn backbone equations summarized in Section~\ref{subsec:radar-baseline}. The backbone first normalizes each instance by z-score,
\begin{equation}
\hat D = \frac{D - \mu(D)}{\sigma(D) + \varepsilon}.
\end{equation}
For the spectral initializer, it uses a rank-\(k\) SVD of the normalized matrix,
\begin{equation}
\hat D \approx U_k \Sigma_k V_k^\top,
\end{equation}
and forms the initial node representation
\begin{equation}
X_{\mathrm{svd}} =
\left[
U_k \Sigma_k^{1/2}
\,\middle\|\,
V_k \Sigma_k^{1/2}
\right],
\qquad
H^0 = W_{\mathrm{proj}}X_{\mathrm{svd}}.
\end{equation}
The left and right singular factors encode source-role and target-role structure before message passing.

At encoder layer \(\ell\), head \(h\) computes a mixed pairwise score
\begin{equation}
\xi_{ij}^{(h)} =
\begin{bmatrix}
\frac{(q_i^{(h)})^\top k_j^{(h)}}{\sqrt{d_h}} \\
\hat D_{ij} \\
\hat D_{ji}
\end{bmatrix},
\qquad
M_{ij}^{(h)}
=
\mathrm{MLP}^{(h)}_{\mathrm{mix}}\!\left(\xi_{ij}^{(h)}\right),
\qquad
A^{(h)} = \mathrm{Sinkhorn}(M^{(h)}).
\end{equation}
Thus the encoder has direct access to both directions of each pairwise cost, while the final decoder score in Eq.~\eqref{eq:base} remains a node-compatibility score.

\begin{table*}[t]
    \centering
    \footnotesize
    \caption{Fixed evaluation files and upstream artifacts used in the controlled SVD/Sinkhorn experiments.}
    \label{tab:app-data-artifacts}
    \resizebox{\textwidth}{!}{%
    \begin{tabular}{llll}
        \toprule
        Task & Fixed evaluation files & Array keys and shapes & Upstream RADAR artifacts \\
        \midrule
        ATSP &
        \path{ATSP100/200/500/1000.npz} in \path{atsp/dataset/} &
        \path{data}: \((1000,n,n)\) &
        \href{https://drive.google.com/file/d/1dwKlConq9AhOObcTu-57wOsxr47FEhHU/view?usp=sharing}{dataset};
        \href{https://drive.google.com/file/d/1vO98NyK3DAaDBAJa5Y6bzWLfyM8QGs_0/view?usp=sharing}{checkpoint} \\
        ACVRP &
        \path{ACVRP100/200/500/1000.npz} in \path{acvrp/dataset/} &
        \path{dist}: \((1000,n+1,n+1)\); \path{demand}: \((1000,n+1)\) &
        \href{https://drive.google.com/file/d/1OdzFHqj_kvaSgHMRO0l4nvEuyr7RV2fK/view?usp=sharing}{dataset};
        \href{https://drive.google.com/file/d/10GFNnGh8pKHZbA-YqkhJj3YdaCEIpzic/view?usp=sharing}{checkpoint} \\
        \bottomrule
    \end{tabular}
    }
\end{table*}

\begin{table*}[t]
    \centering
    \footnotesize
    \caption{Default hyperparameters for the controlled SVD/Sinkhorn experiments.}
    \label{tab:app-hyperparams}
    \begin{tabular}{lll}
        \toprule
        Group & Hyperparameter & Value \\
        \midrule
        Training protocol & Training problem size & 100 \\
        Training protocol & Training seed & 1234 \\
        Training protocol & Epochs / checkpoint for reporting & 2,100 / checkpoint 2100 \\
        Training protocol & Training episodes per epoch & 10,000 \\
        Training protocol & Batch size & 64 \\
        Training protocol & Policy-gradient baseline & POMO baseline \\
        Optimization & Optimizer & Adam \\
        Optimization & Learning rate & $4\times 10^{-4}$ \\
        Optimization & Weight decay & $10^{-6}$ \\
        Optimization & LR decay milestones / factor & epochs 2001 and 2101 / 0.1 \\
        Model & Encoder layers & 5 \\
        Model & Embedding dimension & 256 \\
        Model & Attention heads & 8 \\
        Model & Query/key/value dimension & 32 \\
        Model & Feed-forward hidden dimension & 512 \\
        Model & SVD spectral rank & 10 \\
        Model & Sinkhorn attention type / iterations & uniform Sinkhorn / 10 \\
        Decoder & Edge-bias MLP hidden dimension & 32 \\
        Decoder & Training / evaluation logit clipping & 10 / 50 \\
        Evaluation & Target sizes & 100, 200, 500, 1000 \\
        Evaluation & Held-out test instances per target size & 1,000 \\
        Evaluation & Test-time augmentation & disabled \\
        \bottomrule
    \end{tabular}
\end{table*}

\section{Static vs. Masked Lookahead}
\label{sec:lookahead-mask-comparison}

The lookahead used in the main method is called static lookahead, with the definition in Section~\ref{subsec:edge-aware-decoder}:
\begin{equation}
\phi_{t,\mathrm{static}}^{\mathrm{look}}(j)=
\left[
\min_{\ell\ne j}\hat D_{j\ell},
\min_{\ell\ne j}\hat D_{\ell j}
\right].
\end{equation}
This descriptor can be precomputed because the minimization ranges over the full node set, excluding only the candidate itself. Masked lookahead instead recomputes the same one-edge connectivity quantities over the remaining unvisited set at decoding step \(t\):
\begin{equation}
\phi_{t,\mathrm{masked}}^{\mathrm{look}}(j)=
\left[
\min_{\ell\in U_t\setminus\{j\}}\hat D_{j\ell},
\min_{\ell\in U_t\setminus\{j\}}\hat D_{\ell j}
\right],
\end{equation}
where \(U_t\) denotes the unvisited candidates before selecting \(j\). This variant is closer to a state-dependent continuation proxy, but it requires candidate-specific masked reductions during decoding. Table~\ref{tab:lookahead-mask-results} compares the two choices while keeping the local and closure terms enabled in both rows.

\begin{table*}[t]
    \centering
    \footnotesize
    \caption{Static versus masked lookahead on top of the same local and closure edge terms. Time is reported in minutes. Arrows indicate preferred direction.}
    \label{tab:lookahead-mask-results}
    \resizebox{\textwidth}{!}{%
    \begin{tabular}{lc*{4}{ccc}}
        \toprule
        & & \multicolumn{3}{c}{ATSP-100} & \multicolumn{3}{c}{ATSP-200} & \multicolumn{3}{c}{ATSP-500} & \multicolumn{3}{c}{ATSP-1000} \\
        \cmidrule(lr){3-5} \cmidrule(lr){6-8} \cmidrule(lr){9-11} \cmidrule(lr){12-14}
        Method & Lookahead & Obj$\downarrow$ & Time$\downarrow$ & Gap$\downarrow$ & Obj$\downarrow$ & Time$\downarrow$ & Gap$\downarrow$ & Obj$\downarrow$ & Time$\downarrow$ & Gap$\downarrow$ & Obj$\downarrow$ & Time$\downarrow$ & Gap$\downarrow$ \\
        \midrule
        Ours & Static & 1.5741 & 0.05 & 0.63\% & \textbf{1.5852} & 0.38 & \textbf{0.83\%} & \textbf{1.6032} & 5.31 & \textbf{1.71\%} & \textbf{1.6170} & 43.44 & \textbf{2.73\%} \\
        Ours & Masked & \textbf{1.5736} & 0.13 & \textbf{0.59\%} & 1.5854 & 1.69 & 0.85\% & 1.6040 & 46.07 & 1.76\% & 1.6178 & 844.16 & 2.79\% \\
        \bottomrule
    \end{tabular}
    }
\end{table*}

Masked lookahead slightly improves the near-training-scale ATSP-100 result, reducing the gap from $0.63\%$ to $0.59\%$. This advantage does not carry over to larger sizes: static lookahead is marginally better at ATSP-200, ATSP-500, and ATSP-1000, with gaps of $0.83\%$, $1.71\%$, and $2.73\%$ compared with $0.85\%$, $1.76\%$, and $2.79\%$ for masked lookahead. The runtime difference is much larger than the objective difference, rising from $43.44$ minutes to $844.16$ minutes at ATSP-1000. The result supports using static lookahead as the main design point: it gives a cheap continuation cue, while the more state-specific masked variant does not improve scale-generalization under this benchmark setting.

\section{Exploratory Transfer to POMO on Symmetric TSP}
\label{app:pomo-tsp}

To test whether decision-time score modifications remain useful beyond the main asymmetric-routing setting, we also ran a small transfer study on symmetric Euclidean TSP with a POMO backbone \citep{kwon2020pomo}. This comparison is deliberately kept in the appendix. The task family is different, the main paper's Bellman-style asymmetry interpretation does not transfer literally to symmetric TSP, and the available checkpoints do not form a full homogeneous sweep. The narrower question here is only whether making the final decoder score more transition-aware can still help cross-size generalization in a standard node-based routing backbone.

We report the smaller TSP-20 transfer pilot first, followed by the TSP-100 transfer comparison. Table~\ref{tab:pomo-transfer-tsp20} shows models trained on TSP-20 and evaluated on TSP-20/50/100. With $\times 8$ augmentation, adding Ours leaves TSP-20 nearly unchanged, improves TSP-50 from $0.70\%$ gap to $0.54\%$, and improves TSP-100 from $9.88\%$ to $4.96\%$.

\begin{table*}[t]
    \centering
    \footnotesize
    \caption{Exploratory transfer check on symmetric TSP with a POMO backbone trained on TSP-20 and evaluated on TSP-20/50/100. Gap is measured relative to the Concorde reference row. Time is reported in minutes. Bold marks the best learned value in each metric column.}
    \label{tab:pomo-transfer-tsp20}
    \resizebox{\textwidth}{!}{%
    \begin{tabular}{lccccccccc}
        \toprule
        & \multicolumn{3}{c}{TSP-20} & \multicolumn{3}{c}{TSP-50} & \multicolumn{3}{c}{TSP-100} \\
        \cmidrule(lr){2-4} \cmidrule(lr){5-7} \cmidrule(lr){8-10}
        Method & Obj & Gap & Time & Obj & Gap & Time & Obj & Gap & Time \\
        \midrule
        Concorde & 3.8300 & 0.00\% & 5.00 & 5.6900 & 0.00\% & 13.00 & 7.7600 & 0.00\% & 60.00 \\
        \midrule
        POMO & 3.8316 & 0.04\% & \textbf{1.34} & 5.7808 & 1.60\% & \textbf{4.56} & 8.7882 & 13.25\% & \textbf{21.75} \\
        POMO + aug$\times 8$ & 3.8306 & 0.02\% & \textbf{1.34} & 5.7297 & 0.70\% & \textbf{4.56} & 8.5267 & 9.88\% & \textbf{21.75} \\
        POMO + Ours & 3.8308 & 0.02\% & 2.01 & 5.7596 & 1.22\% & 13.39 & 8.3303 & 7.35\% & 89.40 \\
        POMO + Ours + aug$\times 8$ & \textbf{3.8299} & \textbf{0.00\%} & 2.01 & \textbf{5.7208} & \textbf{0.54\%} & 13.39 & \textbf{8.1451} & \textbf{4.96\%} & 89.40 \\
        \bottomrule
    \end{tabular}
    }
\end{table*}

Table~\ref{tab:pomo-transfer} reports the TSP-100 comparison. We use the epoch-210 POMO checkpoint as the primary baseline, because the original POMO paper reports that performance is already close to convergence around this stage, while extending all decoder-augmented runs to epoch 2000 would be substantially more expensive. For context, we also include the original-paper epoch-2000 POMO checkpoint as a longer-training reference. The pattern is consistent with the main paper's broader empirical message. Longer training helps near the training scale: the epoch-2000 POMO checkpoint is the best learned baseline at TSP-100/200. The decoder-augmented model instead helps mainly at larger scales. Relative to the protocol-matched epoch-210 POMO baseline, the decoder-augmented variant reduces the gap from $15.23\%$ to $7.43\%$ at TSP-500 and from $27.40\%$ to $11.61\%$ at TSP-1000, at a runtime cost of roughly $6.5\times$ and $5.3\times$, respectively. It is also much stronger than the longer-trained epoch-2000 POMO checkpoint on the same large instances, improving the TSP-1000 gap from $40.54\%$ to $11.61\%$.

\begin{table*}[t]
    \centering
    \footnotesize
    \caption{Exploratory transfer check on symmetric TSP with a POMO backbone trained on TSP-100. Gap is measured relative to the Concorde reference row. The epoch-2000 POMO checkpoint is included only as a longer-training reference from the original POMO setting, not as a fully protocol-matched baseline for the decoder-augmented runs. Time is reported in minutes. Bold marks the best learned value in each metric column.}
    \label{tab:pomo-transfer}
    \resizebox{\textwidth}{!}{%
    \begin{tabular}{lcccccccccccc}
        \toprule
        & \multicolumn{3}{c}{TSP-100} & \multicolumn{3}{c}{TSP-200} & \multicolumn{3}{c}{TSP-500} & \multicolumn{3}{c}{TSP-1000} \\
        \cmidrule(lr){2-4} \cmidrule(lr){5-7} \cmidrule(lr){8-10} \cmidrule(lr){11-13}
        Method & Obj & Gap & Time & Obj & Gap & Time & Obj & Gap & Time & Obj & Gap & Time \\
        \midrule
        Concorde & 7.7600 & 0.00\% & 5.00 & 10.6683 & 0.00\% & 15.76 & 16.5500 & 0.00\% & 37.66 & 23.1200 & 0.00\% & 399.00 \\
        \midrule
        POMO + aug$\times 8$ (epoch 210) & 7.8060 & 0.59\% & \textbf{21.23} & 11.0026 & 3.13\% & 144.00 & 19.0705 & 15.23\% & \textbf{160.20} & 29.4544 & 27.40\% & \textbf{160.20} \\
        POMO + Ours + aug$\times 8$ (epoch 210) & 7.7876 & 0.36\% & 89.40 & 10.8985 & 2.16\% & 683.40 & \textbf{17.7800} & \textbf{7.43\%} & 1036.20 & \textbf{25.8044} & \textbf{11.61\%} & 840.60 \\
        POMO + aug$\times 8$ (epoch 2000, original checkpoint) & \textbf{7.7740} & \textbf{0.18\%} & 21.88 & \textbf{10.8569} & \textbf{1.77\%} & \textbf{142.20} & 20.1675 & 21.86\% & 200.39 & 32.4935 & 40.54\% & \textbf{160.20} \\
        \bottomrule
    \end{tabular}
    }
\end{table*}

We therefore view the POMO evidence as a useful boundary check rather than a second main benchmark: it suggests that candidate-specific decision-time cues may help cross-size transfer even in symmetric routing, but it does not by itself support the paper's asymmetry-specific mechanism claim.

\section{ELG-style Decoder Bias on TSPLIB and VRPLIB-X}
\label{app:elg-decoder-bias}

We also tested the same score-level idea in an ELG-style setting \citep{gao2024elg}. This check is intentionally kept in the appendix because it uses different benchmark families and a different solver stack from the main controlled asymmetric experiments. The decoder-bias runs are trained with \texttt{only\_global}, without \texttt{ensemble} or \texttt{distance\_penalty}; the selected best rows in Table~\ref{tab:elg-decoder-bias-summary} use \texttt{distance\_penalty} at test time. The CVRP reference row uses the official ELG checkpoint with \texttt{ensemble}. These protocol differences make the table a boundary check rather than a clean leaderboard comparison.

\begin{table*}[t]
    \centering
    \footnotesize
    \caption{Exploratory ELG-style decoder-bias check on TSPLIB and VRPLIB-X. Lower gap is better; more negative $\Delta$Total is better. The ``Large'' bucket denotes 500--1002 for TSPLIB and $>500$ for VRPLIB-X.}
    \label{tab:elg-decoder-bias-summary}
    \begin{tabular}{llccccc}
        \toprule
        Task & Method & $\leq$200 & 200--500 & Large & Total & $\Delta$Total \\
        \midrule
        TSPLIB & Official ELG & 1.1803\% & \textbf{4.3448\%} & 8.9125\% & 3.1245\% & 0.0000pp \\
        TSPLIB & Ours & \textbf{1.0869\%} & 4.7484\% & \textbf{8.2614\%} & \textbf{3.0832\%} & \textbf{-0.0413pp} \\
        \midrule
        VRPLIB-X & Official ELG & 4.5099\% & \textbf{5.5244\%} & 7.8047\% & 6.0309\% & 0.0000pp \\
        VRPLIB-X & Ours & \textbf{4.4725\%} & 5.6105\% & \textbf{7.4471\%} & \textbf{5.9478\%} & \textbf{-0.0830pp} \\
        \bottomrule
    \end{tabular}
\end{table*}

The pattern is consistent with the POMO boundary check but remains too mixed for a main-paper claim. On TSPLIB, Ours reduces total gap from $3.1245\%$ to $3.0832\%$, with gains in the $\leq 200$ and 500--1002 buckets, but it is worse on 200--500 instances. On VRPLIB-X, Ours reduces total gap from $6.0309\%$ to $5.9478\%$, mainly because the $>500$ bucket improves from $7.8047\%$ to $7.4471\%$; the 200--500 bucket still trails the official reference. The evidence therefore supports a cautious boundary-check statement: decoder-side scoring can interact productively with large-scale inference heuristics, but the current ELG-style results do not isolate the decoder contribution as cleanly as the main ATSP diagnostics.

\section{Directed-transition Diagnostic Protocol}
\label{app:transition-diagnostics}

This appendix gives the protocol behind Section~\ref{sec:transition-diagnostics}. The diagnostic is motivated by the final decoder score in Eq.~\eqref{eq:base}, where candidate \(j\) is scored through compatibility between the decoder context and the candidate node embedding. Although the context is conditioned on the current node \(i=c_t\), the directed edge \(i\rightarrow j\) is not an explicit scalar in the final score. The diagnostic therefore asks whether changes to the directed transition are reflected in the candidate logit after the full encoder--decoder computation.

For each ATSP size and model, we collect 1000 decoding states. Each state records the start node, the current node \(i=c_t\), the visited set, the feasible candidate mask, and the candidate logits to be evaluated. Let \(\ell_t(j;D)\) denote the final decoder logit for candidate \(j\) under distance matrix \(D\). For the perturbation-based diagnostics, we construct a counterfactual matrix \(\widetilde D\), rerun the full encoder and decoder with the same partial route, and compare \(\Delta \ell_j=\ell_t(j;\widetilde D)-\ell_t(j;D)\). This end-to-end rerun allows RADAR to respond through its representation; the diagnostic is not restricted to the explicit edge-aware bias.

The four reported diagnostics in Figure~\ref{fig:transition-diagnostics} test the same decision-time question from related views. Current SignAcc increases \(D_{ij}\), so a transition-sensitive decoder should reduce the logit of \(j\), giving \(\Delta \ell_j<0\). Reverse SignAcc increases \(D_{ji}\); this is a signed-response control rather than a correctness target for action \(i\rightarrow j\), so values near chance are expected. Antisym SignAcc increases \(D_{ij}\) while decreasing \(D_{ji}\), and should again reduce the logit of \(j\) if the decoder is sensitive to the current directed transition. Local PairAcc uses the same recorded decoding states but samples matched feasible candidates whose closure and lookahead quantities are close; it measures whether the candidate with the cheaper current directed edge receives the higher logit.

We also compute a swap control, which exchanges \(D_{ij}\) and \(D_{ji}\). Its expected sign follows the induced change in the current edge, decreasing the logit when the swapped-in \(D_{ji}\) is larger than \(D_{ij}\) and increasing it when it is smaller. Cases with nearly equal directed costs provide weak signed evidence and are treated as control behavior rather than primary benchmark evidence.

\end{document}